\documentclass[twoside,11pt]{article}

\usepackage{blindtext}

\usepackage{jmlr2e}
\usepackage{xcolor}         % colors

\usepackage{lastpage}
\jmlrheading{26}{2025}{1-\pageref{LastPage}}{12/23}{2/25}{23-1634}{Jianqing Zhang, Yang Liu, Yang Hua, Hao Wang, Tao Song, Zhengui Xue, Ruhui Ma, and Jian Cao}

\ShortHeadings{\pfllib: A Personalized Federated Learning Library and Benchmark Platform}{Zhang, Liu, Hua, Wang, Song, Xue, Ma, and Cao}
\firstpageno{1}

\makeatletter
\DeclareRobustCommand\onedot{\futurelet\@let@token\@onedot}
\def\@onedot{\ifx\@let@token.\else.\null\fi\xspace}

\makeatother

\usepackage[capitalize]{cleveref}
\crefname{section}{Sec.}{Secs.}
\Crefname{section}{Section}{Sections}
\Crefname{table}{Table}{Tables}
\crefname{table}{Tab.}{Tabs.}

\usepackage{subcaption}
\usepackage{booktabs}

\usepackage{xspace}
\def\pfllib{\texttt{PFLlib}\xspace}

\definecolor{SJTUred}{RGB}{202, 21, 30}
\newcommand{\linecode}[1]{\colorbox[rgb]{1,1,1}{\color{SJTUred} \texttt{#1}}}

\usepackage{listings}
\usepackage{xcolor}
\definecolor{codegreen}{rgb}{0,0.6,0}
\definecolor{codegray}{rgb}{0.5,0.5,0.5}
\definecolor{codeblue}{RGB}{88,160,211}
\definecolor{backcolour}{RGB}{238,238,238}
\lstdefinestyle{mystyle}{
    backgroundcolor=\color{backcolour},   
    commentstyle=\color{codegreen},
    keywordstyle=\color{codeblue},
    numberstyle=\tiny\color{codegray},
    stringstyle=\color{codegreen},
    basicstyle=\ttfamily\footnotesize,
    breakatwhitespace=false,         
    breaklines=true,                 
    captionpos=b,                    
    keepspaces=true,                 
    numbers=left,                    
    numbersep=5pt,                  
    showspaces=false,                
    showstringspaces=false,
    showtabs=false,                  
    tabsize=2
}
\usepackage{multirow}
\usepackage{array}

\hypersetup{hidelinks}

\begin{document}

\title{\pfllib: A Beginner-Friendly and Comprehensive Personalized Federated Learning Library and Benchmark}

\author{\name Jianqing Zhang$^1$ \email tsingz@sjtu.edu.cn \\
\name Yang Liu$^{2,*}$ \email liuy03@air.tsinghua.edu.cn \\
\name Yang Hua$^3$ \email y.hua@qub.ac.uk \\
\name Hao Wang$^4$ \email hwang9@stevens.edu \\
\name Tao Song$^1$ \email songt333@sjtu.edu.cn \\
\name Zhengui Xue$^1$ \email zhenguixue@sjtu.edu.cn \\
\name Ruhui Ma$^1$ \email ruhuima@sjtu.edu.cn \\
\name Jian Cao$^{1,*}$ \email cao-jian@sjtu.edu.cn \\
\addr $^1$ Shanghai Jiao Tong University, Shanghai, China\\
\addr $^2$ Institute for AI Industry Research, Tsinghua University, Beijing, China\\
\addr $^3$ Queen's University Belfast, Belfast, UK\\
\addr $^4$ Stevens Institute of Technology, New Jersey, USA\\
\addr $^*$ Corresponding Authors
}

\editor{Albert Bifet}

\maketitle

\begin{abstract}%   <- trailing '%' for backward compatibility of .sty file
Amid the ongoing advancements in Federated Learning (FL), a machine learning paradigm that allows collaborative learning with data privacy protection, personalized FL (pFL) has gained significant prominence as a research direction within the FL domain. Whereas traditional FL (tFL) focuses on jointly learning a global model, pFL aims to balance each client's global and personalized goals in FL settings. To foster the pFL research community, we started and built \pfllib, a comprehensive pFL library with an integrated benchmark platform. In \pfllib, we implemented 37 state-of-the-art FL algorithms (8 tFL algorithms and 29 pFL algorithms) and provided various evaluation environments with three statistically heterogeneous scenarios and 24 datasets. 
At present, \pfllib\footnote{\url{https://www.pfllib.com/} and \url{https://github.com/TsingZ0/PFLlib}} has gained more than 1600 stars and 300 forks on GitHub. 
\end{abstract}

\begin{keywords}
  federated learning, personalization, privacy, benchmark, heterogeneity
\end{keywords}

\section{Introduction}

Federated Learning (FL) has gained significant attention due to its ability to perform distributed machine learning while ensuring privacy preservation~\citep{yang2019federated}. In traditional FL (tFL) algorithms, such as FedAvg~\citep{mcmahan2017communication}, participating clients train local models using local data and send only local model updates to a global server, which then aggregates these updates to obtain a global model. These approaches do not consider the customization needs of each local client. %However, in practice, the Artificial Intelligence (AI) model trained collaboratively using traditional FL (tFL) algorithms, such as FedAvg~\citep{mcmahan2017communication}, faces the challenge of statistical heterogeneity. This includes issues such as not non-independent and identically distributed (Non-IID) data among the participating clients. Next, 
Personalized FL (pFL) is introduced to train customized client models to improve their performance on individualized tasks. %unlike tFL approaches that only train a single global model. 
In tandem with the burgeoning prominence of pFL, there has been a surge in the development of various pFL algorithms and associated techniques~\citep{tan2022towards, zhang2023gpfl}. However, due to their rapid progress and diverse settings, the difficulties of tracking, implementing, and benchmarking these methods also grow tremendously. 

To alleviate these challenges, we have developed \pfllib, a comprehensive pFL library with an integrated benchmark platform. \pfllib includes implementations of \textbf{37} state-of-the-art (SOTA) tFL and pFL algorithms, encompassing 8 tFL algorithms and \textbf{29} pFL algorithms. Our library is beginner-friendly and easily extensible, allowing contributors to seamlessly add new algorithms, scenarios, and datasets, thus ensuring that \pfllib remains up-to-date and popular. In addition, we have implemented three types of data heterogeneity scenarios and incorporated 24 datasets, covering Computer Vision (CV), Natural Language Processing (NLP), and Sensor Signal Processing (SSP) tasks. We can evaluate FL algorithms in \pfllib and assess their adaptability to various scenarios, providing valuable information for algorithm selection and evaluation in practical applications. 

\section{Related Work}

With the rapid development of the FL field, numerous benchmarks and platforms have emerged in recent years. Most of their latest versions are for practical deployments, such as FATE~\citep{liu2021fate}, FedML~\citep{he2020fedml}, 
FederatedScope~\citep{federatedscope}, Flower~\citep{beutel2020flower}, TensorFlow Federated\footnote{\url{https://www.tensorflow.org/federated}}, NVIDIA Clara\footnote{\url{https://developer.nvidia.com/industries/healthcare}}, SecretFlow\footnote{\url{https://github.com/secretflow/secretflow}}, Fedlearner\footnote{\url{https://github.com/bytedance/fedlearner}}, and
PySyft\footnote{\url{https://github.com/OpenMined/PySyft}}. 
% \yang{many of these projects are not "built for industrial applications" only, I would argue, such as FedML, PySyft, Tensorflow TF. I think it is better to stress that existing projects lack coverage on pFL algorithms, followed by a detailed discussion on pFL coverage in existing FL projects. I.e., are there pFL projects? How to compare ours against them in terms of coverage, implementation etc.}
Despite the efficient resource management and extensive functionality offered by these platforms, they can present a challenge for beginners who seek to comprehend the fundamental mechanisms of FL and delve into the philosophical aspects of existing FL algorithms. There are also some beginner-friendly platforms, such as 
LEAF~\citep{caldas2018leaf}, NIID-Bench~\citep{li2022federated}, Motley~\citep{wu2022motley}, OARF~\citep{hu2022oarf}, FedEval~\citep{chai2020fedeval}, and FedLab~\citep{zeng2023fedlab}.
However, these benchmarks and platforms still lack sufficient and up-to-date built-in SOTA pFL algorithms for researchers to learn, compare, and analyze. 
% \yang{need to be more detailed in comparison. I remember we discussed some recent PFL projects which are accepted by major conferences, are they all included?}

pFL-Bench~\citep{chen2022pfl} is one of the latest PFL projects, %which includes 20 pFL methods. However, there are 
but it supports only 5 SOTA pFL methods while the remaining pFL methods are variants created by combining them with existing approaches including FedBN~\citep{li2020fedbn}, FedOpt~\citep{asad2020fedopt}, and Fine-tuning (FT). Besides, all the pFL algorithms in pFL-Bench are outdated (before 2022). 
% Besides, it is also difficult for pFL-Bench to classify pFL methods with limited SOTA pFL methods. 
In contrast, our \pfllib consists of 29 SOTA pFL algorithms. Moreover, due to our straightforward file structure, \pfllib is more accessible for beginners to learn and utilize pFL algorithms compared to the complex pFL-Bench. 
% \yang{any other differences/advantages of ours compared to pFL-Bench worth mentioning? In terms of how to use, extend, and evaluate maybe? e.g., metrics used to evaluate, efficiency, privacy etc. }
% We also classify these 29 pFL algorithms into 6 categories. 

\section{\pfllib: A Beginner-Friendly and Comprehensive Library}

\noindent\textbf{Algorithms.} 
In our \pfllib, the primary focus is on pFL algorithms. In addition, we have also included a selection of tFL algorithms to facilitate the evaluation of pFL algorithms, following previous pFL research ~\citep{li2021ditto, t2020personalized, zhang2023fedala}. Based on their foundational techniques, we have categorized 8 tFL algorithms and 29 pFL algorithms. The detailed classification is presented in \Cref{tab:algo}. 

\begin{table}[h]
  \centering
  \caption{FL Algorithm Taxonomy in our \pfllib.}
  \resizebox{\linewidth}{!}{
    \begin{tabular}{l|l|p{105mm}}
    \toprule
     & Category & Algorithms \\
    \midrule
    \parbox[t]{2mm}{\multirow{6}{*}{\rotatebox[origin=c]{90}{8 tFL Algorithms}}} & Basic tFL & FedAvg~\citep{mcmahan2017communication} \\
    \cmidrule{2-3}
    & Update-correction-based tFL & SCAFFOLD~\citep{karimireddy2020scaffold} \\
    \cmidrule{2-3}
    & Regularization-based tFL & FedProx~\citep{MLSYS2020_38af8613} and FedDyn~\citep{acarfederated} \\
    \cmidrule{2-3}
    & Model-splitting-based tFL & MOON~\citep{li2021model} and FedLC~\citep{zhang2022federated1} \\
    \cmidrule{2-3}
    & Knowledge-distillation-based tFL & FedGen~\citep{zhu2021data} and FedNTD~\citep{lee2022preservation} \\
    \midrule
    \parbox[t]{2mm}{\multirow{15}{*}{\rotatebox[origin=c]{90}{29 pFL Algorithms}}} & Meta-learning-based pFL & Per-FedAvg~\citep{NEURIPS2020_24389bfe} \\
    \cmidrule{2-3}
    & Regularization-based pFL & pFedMe~\citep{t2020personalized} and Ditto~\citep{li2021ditto} \\
    \cmidrule{2-3}
    & \multirow{3}{*}{Personalized-aggregation-based pFL} & APFL~\citep{deng2020adaptive}, FedFomo~\citep{zhang2020personalized}, FedAMP~\citep{huang2021personalized}, FedPHP~\citep{li2021fedphp}, APPLE~\citep{ijcai2022p301}, and FedALA~\citep{zhang2023fedala} \\
    \cmidrule{2-3}
    & \multirow{5}{*}{Model-splitting-based pFL} & FedPer~\citep{arivazhagan2019federated}, LG-FedAvg~\citep{liang2020think}, FedRep~\citep{collins2021exploiting}, FedRoD~\citep{chen2021bridging}, FedBABU~\citep{ohfedbabu}, FedGC~\citep{niu2022federated}, FedCP~\citep{Zhang2023fedcp}, GPFL~\citep{zhang2023gpfl}, FedGH~\citep{yi2023fedgh}, DBE~\citep{zhang2023eliminating}, FedCAC~\citep{wu2023bold}, and PFL-DA~\citep{shi2023personalized} \\
    \cmidrule{2-3}
    & \multirow{3}{*}{Knowledge-distillation-based pFL} & FedDistill~\citep{seo202216}, FML~\citep{shen2020federated}, FedKD~\citep{wu2022communication}, FedProto~\citep{tan2022fedproto}, FedPCL~\citep{tan2022federated}, and FedPAC~\citep{xu2022personalized} \\    
    \cmidrule{2-3}
    & Other pFL & FedMTL~\citep{seo202216} and FedBN~\citep{li2020fedbn} \\  
    \bottomrule
    \end{tabular}}
    \label{tab:algo}
\end{table}

\noindent\textbf{Scenarios and Datasets.} 
%According to the popular survey~\citep{li2022federated}, there are mainly three kinds of statistical heterogeneity, including the label distribution skew, the feature distribution skew, and the quantity skew. 
%we mix the quantity skew into the label distribution skew and the feature distribution skew, then we call these 
In \pfllib, we first consider two types of scenarios for data heterogeneity: \textit{label skew} and \textit{feature shift}, where different client datasets differ in label categories and feature categories, respectively~\citep{zhang2023gpfl}. Both CV and NLP classification tasks are considered in these two scenarios. Moreover, we also introduce a \textit{real world} scenario with naturally collected datasets from distributed sensors (HAR~\citep{anguita2012human} and PAMAP2~\citep{reiss2012introducing} for SSP tasks), hospitals (Camelyon17~\citep{koh2021wilds} for CV tasks), and camera traps (iWildCam~\citep{koh2021wilds} for CV tasks) to evaluate algorithms' performance in realistic scenarios. 
% Specifically, we assign the data collected from each single entity (\eg, a sensor or hospital) to each client.

\noindent\textbf{Privacy Evaluation.} We implement the popular Deep Leakage from Gradients (DLG) attack~\citep{zhu2019deep} and the Peak Signal-to-Noise Ratio (PSNR) metric~\citep{wufedcg} to evaluate the privacy-preserving abilities of existing tFL/pFL algorithms. 

\noindent\textbf{Easy to Use and Extend.} 
In \pfllib, algorithms are implemented by \textbf{three} critical files: \linecode{serverX.py} for server creation, \linecode{clientX.py} for client creation, and \linecode{main.py} for hyperparameter configuration. Here, ``\linecode{X}'' represents some algorithm name. We can simply create a new algorithm \linecode{Y} by only adding specific features to \linecode{serverY.py} and \linecode{clientY.py} and while utilizing the core APIs in \linecode{serverbase.py} and \linecode{clientbase.py}. 
To create a scenario, users simply need \textbf{one} command line and run the evaluation with another \textbf{one} command line. 
% \yang{add exactly what are these command lines. maybe use a figure.}
We show an example of using FedALA on MNIST~\citep{lecun1998gradient}:
\begin{lstlisting}[language=Bash]
# generate a practical non-iid and unbalanced scenario using MNIST
python generate_MNIST.py noniid - dir # in ./dataset
# evaluate the FedALA algorithm using a CNN with default hyperparameters
python main.py -data MNIST -m CNN -algo FedALA -gr 2000 -did 0 # in ./system
\end{lstlisting}

\noindent\textbf{Impacts.} 
Our \pfllib is active and popular in the pFL community, as shown by the increasing number of GitHub stars, forks, and active discussions. Due to its simplicity and extensibility, numerous new platforms and projects have been built upon it, such as the FL-bench\footnote{\url{https://github.com/KarhouTam/FL-bench/tree/c11efc286dab4565245da34d7300d5bb07b87a0a}}, the HtFLlib\footnote{\url{https://github.com/TsingZ0/HtFLlib}}, and the FL-IoT\footnote{\url{https://github.com/TsingZ0/FL-IoT}}. Besides, the experiments of several latest SOTA methods~\citep{zhang2023fedala, Zhang2023fedcp, zhang2023gpfl, zhang2023eliminating, zhang2024fedtgp, zhang2024upload, zhang2024fedl2g} are also conducted using our \pfllib. 

\noindent\textbf{Benchmark.} 
Due to limited space here, we only evaluate 20 algorithms in two \textit{label skew} scenarios following the default settings\footnote{Due to frequent updates, some default settings and codes for scenario creation may change in \pfllib.} of GPFL~\citep{zhang2023gpfl}. Please refer to our official website\footnote{\url{https://www.pfllib.com/}} for more documents, details, and results. In \Cref{tab:label_skew}, we use the 4-layer CNN~\citep{mcmahan2017communication} for CV tasks on Fashion-MNIST (FMNIST)~\citep{xiao2017fashion}, Cifar100, and Tiny-ImageNet~\citep{chrabaszcz2017downsampled} (TINY for short) datasets and use the fastText~\citep{joulinetal2017bag} for NLP tasks on AG News~\citep{zhang2015character} dataset. We also use ResNet-18~\citep{he2016deep} on Tiny-ImageNet and denote it TINY*.

\begin{table*}[ht]
  \centering
  \caption{The test accuracy (\%) on the CV and NLP tasks in \textit{label skew} settings. }
  \resizebox{\linewidth}{!}{
    \begin{tabular}{l|ccc|cccccc}
    \toprule
    \textbf{Settings} & \multicolumn{3}{c|}{\textbf{Pathological \textit{Label Skew} Setting}} & \multicolumn{5}{c}{\textbf{Practical \textit{Label Skew} Setting}} \\
    \midrule
     & FMNIST & Cifar100 & TINY & FMNIST & Cifar100 & TINY & TINY* & AG News\\
    \midrule
    \textbf{FedAvg} & 80.41$\pm$0.08 & 25.98$\pm$0.13 & 14.20$\pm$0.47 & 85.85$\pm$0.19 & 31.89$\pm$0.47 & 19.46$\pm$0.20 & 19.45$\pm$0.13 & 87.12$\pm$0.19 \\
    \textbf{FedProx} & 78.08$\pm$0.15 & 25.94$\pm$0.16 & 13.85$\pm$0.25 & 85.63$\pm$0.57 & 31.99$\pm$0.41 & 19.37$\pm$0.22 & 19.27$\pm$0.23 & 87.21$\pm$0.13 \\
    % \textbf{MOON} & 70.67$\pm$0.38  & 25.65$\pm$0.21 & 14.93$\pm$0.42 & 83.69$\pm$0.52 & 32.37$\pm$0.48 & 19.68$\pm$0.20 & 19.02$\pm$0.21 & 84.14$\pm$0.17\\
    \textbf{FedGen}& 79.76$\pm$0.60 & 20.80$\pm$1.00 &13.82$\pm$0.09 & 84.90$\pm$0.31 & 30.96$\pm$0.54 & 19.39$\pm$0.18 & 18.53$\pm$0.32 & 89.86$\pm$0.83\\
    \midrule
    \textbf{Per-FedAvg} & 99.18$\pm$0.08 & 56.80$\pm$0.26 & 28.06$\pm$0.40 & 95.10$\pm$0.10 & 44.28$\pm$0.33 & 25.07$\pm$0.07 & 21.81$\pm$0.54 & 87.08$\pm$0.26 \\
    \midrule
    \textbf{pFedMe} & 99.35$\pm$0.14 & 58.20$\pm$0.14 & 27.71$\pm$0.40 & 97.25$\pm$0.17 & 47.34$\pm$0.46 & 26.93$\pm$0.19 & 33.44$\pm$0.33 & 87.08$\pm$0.18 \\
    \textbf{Ditto} & 99.44$\pm$0.06 & 67.23$\pm$0.07 & 39.90$\pm$0.42 & 97.47$\pm$0.04 & 52.87$\pm$0.64 & 32.15$\pm$0.04 & 35.92$\pm$0.43 & 91.89$\pm$0.17 \\
    \midrule
    \textbf{APFL} & 99.41$\pm$0.02 & 64.26$\pm$0.13 & 36.47$\pm$0.44 & 97.25$\pm$0.08 & 46.74$\pm$0.60 & 34.86$\pm$0.43 & 35.81$\pm$0.37 & 89.37$\pm$0.86\\
    \textbf{FedFomo} & 99.46$\pm$0.01  & 62.49$\pm$0.22 & 36.55$\pm$0.50 & 97.21$\pm$0.02 & 45.39$\pm$0.45 & 26.33$\pm$0.22 & 26.84$\pm$0.11 & 91.20$\pm$0.18 \\
    \textbf{FedAMP} &99.42$\pm$0.03  & 64.34$\pm$0.37 &  36.12$\pm$0.30 &97.20$\pm$0.06 & 47.69$\pm$0.49 & 27.99$\pm$0.11 & 29.11$\pm$0.15 & 83.35$\pm$0.05 \\
    \textbf{APPLE} & 99.30$\pm$0.01 & 65.80$\pm$0.08 &  36.22$\pm$0.40 & 97.06$\pm$0.07 & 53.22$\pm$0.20 & 35.04$\pm$0.47 & 39.93$\pm$0.52 & 84.10$\pm$0.18 \\
    \textbf{FedALA} & 99.57$\pm$0.01 & 67.83$\pm$0.06 & 40.31$\pm$0.30 & 97.66$\pm$0.02 & 55.92$\pm$0.03 & 40.54$\pm$0.02 & 41.94$\pm$0.02 & 92.45$\pm$0.10 \\
    \midrule
    \textbf{FedPer} & 99.47$\pm$0.03 & 63.53$\pm$0.21 & 39.80$\pm$0.39 & 97.44$\pm$0.06 & 49.63$\pm$0.54 & 33.84$\pm$0.34 & 38.45$\pm$0.85 & 91.85$\pm$0.24 \\
    \textbf{FedRep} & 99.56$\pm$0.03 & 67.56$\pm$0.31 & 40.85$\pm$0.37 & 97.56$\pm$0.04 & 52.39$\pm$0.35 & 37.27$\pm$0.20 & 39.95$\pm$0.61 & 92.25$\pm$0.20 \\
    \textbf{FedRoD} & 99.52$\pm$0.05 & 62.30$\pm$0.02 & 37.95$\pm$0.22 & 97.52$\pm$0.04 & 50.94$\pm$0.11 & 36.43$\pm$0.05 & 37.99$\pm$0.26 & 92.16$\pm$0.12 \\
    \textbf{FedBABU} & 99.41$\pm$0.05 & 66.85$\pm$0.07 & 40.72$\pm$0.64 & 97.46$\pm$0.07 & 55.02$\pm$0.33 & 36.82$\pm$0.45 & 34.50$\pm$0.62 & 95.86$\pm$0.41\\
    \textbf{FedCP} & 99.66$\pm$0.04 & 71.80$\pm$0.16 & 44.52$\pm$0.22 & \textbf{97.89$\pm$0.05} & 59.56$\pm$0.08 & \textbf{43.49$\pm$0.04} & \textbf{44.18$\pm$0.21} & 92.89$\pm$0.10 \\
    \textbf{GPFL} & \textbf{99.85$\pm$0.08} & 71.78$\pm$0.26 & \textbf{44.58$\pm$0.06} & 97.81$\pm$0.09 & 61.86$\pm$0.31 & 43.37$\pm$0.53 & 43.70$\pm$0.44 & \textbf{97.97$\pm$0.14} \\
    \textbf{FedDBE} & 99.74$\pm$0.04 & \textbf{73.38$\pm$0.18} & 42.89$\pm$0.29 & 97.69$\pm$0.05 & \textbf{64.39$\pm$0.27} & 43.32$\pm$0.37 & 42.98$\pm$0.52 & 96.87$\pm$0.18 \\
    \midrule
    \textbf{FedDistill} & 99.51$\pm$0.03& 66.78$\pm$0.15 & 37.21$\pm$0.25 & 97.43$\pm$0.04 & 49.93$\pm$0.23 & 30.02$\pm$0.09 & 29.88$\pm$0.41 & 85.76$\pm$0.09 \\
    \textbf{FedProto} & 99.49$\pm$0.04 & 69.18$\pm$0.03 & 36.78$\pm$0.07 & 97.40$\pm$0.02 & 52.70$\pm$0.33 & 31.21$\pm$0.16 & 26.38$\pm$0.40 & 96.34$\pm$0.58 \\
    \bottomrule
    \end{tabular}}
  \label{tab:label_skew}
\end{table*}

\vspace{-5mm}

\section{Conclusion}

To support the rapidly evolving pFL research community, we built \pfllib, a beginner-friendly library that includes 37 cutting-edge tFL / pFL algorithms. Besides, we also built a benchmark platform in \pfllib with comprehensive features, datasets, and scenarios. 
% We are committed to the ongoing development of \pfllib, expanding its range of tFL/pFL algorithms, scenarios, datasets, \etc.

% \acks{All acknowledgements go at the end of the paper before appendices and references.
% Moreover, you are required to declare funding (financial activities supporting the
% submitted work) and competing interests (related financial activities outside the submitted work).
% More information about this disclosure can be found on the JMLR website.}
\acks{This work was supported by the National Key R\&D Program of China under Grant No.2022ZD0160504, the Program of Technology Innovation of the Science and Technology Commission of Shanghai Municipality (Granted No. 21511104700), the Interdisciplinary Program of Shanghai Jiao Tong University (project number YG2024QNB05), Tsinghua University(AIR)-Asiainfo Technologies (China) Inc. Joint Research Center, and the National Natural Science Foundation of China (62472284).}

% Manual newpage inserted to improve layout of sample file - not
% needed in general before appendices/bibliography.

\vskip 0.2in
\bibliography{sample}

\end{document}